# VAD-Net: Multidimensional Facial Expression Recognition in Intelligent Education System

Huo Yi
Department of Educational Information Technology
Teachers' College, Beijing Union University
Beijing, China
sfthuoyi@buu.edu.cn

Ge Yun
Department of Computer Teaching and Research
University of Chinese Academy of Social Sciences
Beijing, China
gaiyun@ucass.edu.cn

*Abstract*— Current FER (Facial Expression Recognition) dataset is mostly labeled by emotion categories, such as happy, angry, sad, fear, disgust, surprise, and neutral which are limited in expressiveness. However, future affective computing requires more comprehensive and precise emotion metrics which could be measured by VAD(Valence-Arousal-Dominance) multidimension parameters. To address this, AffectNet has tried to add VA (Valence and Arousal) information, but still lacks D(Dominance). Thus, the research introduces VAD annotation on FER2013 dataset, takes the initiative to label D(Dominance) dimension. Then, to further improve network capacity, it enforces orthogonalized convolution on it, which extracts more diverse and expressive features and will finally increase the prediction accuracy. Experiment results show that D dimension could be measured but is difficult to obtain compared with V and A dimension no matter in manual annotation or regression network prediction. Secondly, the ablation test by introducing orthogonal convolution verifies that better VAD prediction could be obtained in the configuration of orthogonal convolution. Therefore, the research provides an initiative labelling for D dimension on FER dataset, and proposes a better prediction network for VAD prediction through orthogonal convolution. The newly built VAD annotated FER2013 dataset could act as a benchmark to measure VAD multidimensional emotions, while the orthogonalized regression network based on ResNet could act as the facial expression recognition baseline for VAD emotion prediction. The newly labeled dataset and implementation code is publicly available on https://github.com/YeeHoran/VAD-Net .

## I. Introduction

With the rapid development of artificial intelligence and big data technology, intelligent online education appears by integrating intelligence with traditional education and is very popular at present [1]. It provides ubiquitous learning environment that anyone, at any time, in anywhere, can learn anything online with any edge terminal devices (5A) [2]. This promotes knowledge dissemination, distributes high-quality educational resources more efficiently, and contributes a lot to high quality and equal education [3] [4]. What is more, the sudden entry of the global epidemic of COVID-19 has become a catalyst for online education [5][6][7][8], makes it has no choice but to develop rapidly. Furthermore, the intelligence of the online learning system is also required because the students not only want to obtain knowledge, but also to gain more emotion interaction as they did in traditional classes [9]. They hope to experience a consistent emotion interaction as they do in real traditional classroom environments. Thus, personalized learning and adaptive instruction gradually progresses to smart online system and requires intelligent emotion interaction [10][11].

For smart online learning system, it is crucial to add intelligent FER (Facial Expression Recognition) tools for recognizing students' learning emotions in class. Through analyzing students' facial expression features in learning environment, common FER tools detect students' emotion categories in real time and send them to teachers. After obtaining the emotion conditions of students, teachers can adjust their instruction accordingly [12][13]. Thus, the performance of this FER system is critical in promoting the effect of intelligent online education [14] [15].

At present, there are mainly two methods for emotion quantification. The first is discrete emotion categories, such as seven standard emotions: anger, disgust, tension, joy, sadness, surprise, and neutrality. The second is continuous value quantification from multiple dimensions. Mollahosseini [16] expresses emotions in the two-dimensional space of valence (V) and arousal (A) in continuous form. However, Mehrabian [17] found that valence (V) and arousal (A) alone cannot fully express emotional categories. He proposed adding dominance (D, dominance-submissiveness) and demonstrated through research that adopting the VAD three dimensions can effectively explain human emotions, providing a comprehensive quantification and expression of human emotions [18]. In fact, the aforementioned discrete emotion categories all lie in the VAD three-dimensional space, making the latter a more comprehensive and accurate method of emotion quantification. However, current systems almost exclusively use discrete emotion categories for quantification [19], overlooking the analysis from the Valence (V), Arousal (A), Dominance (D) multidimensional perspective. Future smart educational environments need to perceive emotions from multiple angles. Therefore, current emotion quantification methods are challenging to meet the advanced requirements of emotion computation in future intelligent education systems.

The research is supported by China Ministry of Education, Humanities, and social science research projects (No. 23YJE880001).

There is a problem in convolutional computations where feature redundancy affects the performance of the network. Currently, emotion recognition networks face issues of high correlation and redundancy in the extracted features, thereby impacting the model's recognition performance. Existing research on emotion recognition models has not yet explored the network's capability from the perspective of reducing inter-feature correlation and redundancy to enhance recognition accuracy.

This study establishes a novel algorithm with enhanced feature distinctiveness to improve the stability of convolutional neural networks and enhance the accuracy of the network model in emotion recognition.

In conclusion, there are three contributions of the research as below:

- Initiate to label the D dimension emotion parameter on FER dataset. The new annotated VAD FER dataset could act as a benchmark to measure VAD multidimensional emotions from facial expression images.

- Propose a better convolution regression network for VAD prediction through orthogonalized convolution. This ResNet based network could be the baseline for obtaining VAD multidimensional emotion parameters.

- Provide a VAD (Valence, Arousal and Dominance) recognition tool of FER (Facial Expression Recognition) in intelligent emotion interaction.

## II. Background and related works

### A. Affective Computing And Emotion Interaction In Education System

The idea of "enabling computers to recognize user emotions" was initially proposed by the father of artificial intelligence, Professor Minsky from the Massachusetts Institute of Technology [20]. He emphasized that emotions are a crucial factor in achieving computer intelligence. Subsequently, Professor Picard explicitly defined the concept of affective computing in 1997 [21], stating that affective computing involves the computation of factors related to emotions, triggered by emotions, or capable of influencing and determining emotional changes. Later, neuroscientists discovered the significant role of emotions in human cognition and decision-making.

As a result, Pekrun and others provided a definition for learning emotions, which are emotions closely related to school learning, classroom teaching activities, and learning achievements [22][23]. They proposed that learners can experience joy due to smooth learning and anxiety when facing unresolved difficulties in learning.

Li [24] believes that academic emotions have a significant impact on learning outcomes and that emotional data related to learning can be obtained through multiple channels. The literature further suggests that emotional data can be recognized through students' facial expressions, speech, and behavior, with facial expressions being the most reflective of emotional states. Additionally, emotional data in learning can be gathered from various channels, such as facial expression images, physiological data, text data, among others. Facial expression images, in particular, are non-invasive, natural, objective, and best reflect a person's emotional state.

### B. Emotion Measurement Metrics

According to different application contexts, there are various ways to categorize emotional states, such as the classification into 7 standard emotions: anger, disgust, tension, joy, sadness, surprise [22]. Another example is the classification into 20 basic emotions, including boredom, anger, anxiety, surprise, sadness, discouragement, pride, hope, confusion, happiness [23]. In the context of learning, literature [23] defines 5 emotions: enjoyment, confusion, distraction, fatigue, and neutrality. Li [24] also proposes categories for learning emotions, including confusion, curiosity, distraction, enjoyment, fatigue, depression, and neutrality, totaling 7 categories.

Mollahosseini [16] expresses emotional categories in the two-dimensional space of pleasure (V, Valence) and arousal (A, Arousal) in the form of continuous values. Here, V can be quantified into various emotional categories as mentioned earlier, while A can be used to represent the student's level of engagement. Furthermore, this study employs two deep learning models to solve for pleasure (V, Valence) and arousal (A, Arousal). In comparison to the emotional categories mentioned earlier, this method explores emotional values from more perspectives and with finer granularity.

However, Mehrabian and Russell [25] found that pleasure (V) and arousal (A) alone cannot fully express emotional categories. In 1974, they proposed that emotions should have three dimensions: pleasure-displeasure (valence), arousal-nonarousal (arousal), and dominance-submissiveness (control). Pleasure (P) represents the positivity or negativity of the subject's emotional state, indicating the degree of emotional positivity or negativity, reflecting the essence of emotions. Arousal (A) is the level of physiological activation in the subject's nervous system, related to the activation level of bodily energy. Dominance (D) is the subject's control state over the scene and others, used to explain whether the emotional state is subjectively generated by the individual or influenced by the objective environment. Mehrabian demonstrated through research that adopting these three dimensions (PAD) can effectively explain human emotions, providing a comprehensive quantification and expression of human emotions [26].

In fact, AffectNet[27] has pioneered to attach VA labels on FER dataset. Dominance is the most difficult to annotate, since the face image may not show his/her genuine social status or characteristic explicitly. However, in affective computing of teaching scenario, students' understanding on the learning material, and the state of knowledge and skills reserves are related to dominance.

The various works on emotion recognition above have played a crucial inspirational role in the framework of the emotion recognition system in our study. Thus, it provides VAD (Valence, Arousal, Dominance) multi-dimensional and fine-grained emotion quantification dataset and propose a regression network baseline for VAD prediction in emotion interaction system.

*C. Convolution Regression Network*

Since the VAD metrics are real values, the regression model should be used to predict the VAD continuous values. CNN (Convolutional Neural Network) is a deep learning model which simulates the learning process of human brain, establishes a multi-hidden-layer interconnected network directly from the input to the output neurons. Through training the network with large amounts of data, the connection coefficients in the network are adjusted, gradually abstracting the feature extraction process from edge features, local features to holistic features to obtain recognition results [28]. However, the recognition performance of CNN may still be reduced due to various factors which lower the accuracy of recognition.

Firstly, differences between individuals can lower the performance of emotion recognition algorithms. Zhang [29], from the perspective of ignoring identity features, models facial expressions as a subtraction operation from identity-related feature vectors and then recognizes expressions that are unrelated to identity. Gong [30] developed a learning algorithm where facial images of each group are averaged, reducing biases during recognition by using adaptive convolution kernels and attention mechanisms. This approach provides kernel masks and attention channels for each group, activating different regions during recognition, thereby improving facial recognition accuracy for various groups.

Furthermore, the training instability and feature redundancy of deep convolutional neural networks hinder further performance improvement. A promising solution is to impose orthogonality on convolutional filters [31][32]. In [31], the convolution process is viewed as the multiplication of the convolution kernel matrix and the image matrix. By penalizing the difference between the Gram matrix of the convolutional kernel matrix and the identity matrix, orthogonality is enforced on the convolutional kernel matrix. However, the spectral distribution of the obtained convolutional kernel matrix is still non-uniform. Taking it a step further, [32] represents the convolutional kernel as a double-block Toeplitz matrix and imposes orthogonality on this convolution matrix. This presents an effective method for enforcing orthogonality filter on convolutional layer. This method allows for the learning of more diverse and expressive features, stabilizing the spectral distribution of the convolution matrix and providing better robustness and generalization.

## III. VAD (Valence, Arousal and Dominance) Annotation

*A. Multi-Dimensional Emotion (Valence, Arousal & Dominance) Annotation*

The definition of valence, arousal and dominance dimensions was adapted from [3] as below, and was given to annotators in our instructions:

In psychology, Valence, Arousal, and Dominance are dimensions of emotion commonly referred to as the VAD model. These dimensions are used to describe and categorize emotional experiences:

- Valence:

Definition: Valence refers to the positive or negative quality of an emotional experience. It represents the degree of pleasantness or unpleasantness associated with an emotion.

Scale: Positive valence indicates positive or pleasant emotions, while negative valence indicates negative or unpleasant emotions. The scale ranges from highly positive to highly negative.

- Arousal:

Definition: Arousal refers to the level of activation, energy, or alertness associated with an emotional state. It measures the intensity or arousal level of an emotion.

Scale: High arousal corresponds to intense and aroused states, such as excitement or fear, while low arousal corresponds to calm or relaxed states. The scale ranges from low arousal to high arousal.

- Dominance:

Definition: Dominance refers to the sense of control or power associated with an emotional experience. It reflects the perceived control over the situation or the level of influence an emotion has.

Scale: High dominance indicates a feeling of control, mastery, or influence, while low dominance indicates a sense of being controlled or influenced by external factors. The scale ranges from low dominance to high dominance.

These three dimensions are often used together to create a comprehensive representation of emotions. The VAD model is widely employed in psychology and affective computing to analyze and understand the emotional content of various stimuli, including images, videos, and text. It provides a more nuanced and detailed framework for describing emotions beyond simple category labels.

This dimensional model represents a three-dimensional space (xyz), where V corresponds to the x-axis, A to the y-axis, and D to the z-axis. To simplify annotation, each dimension is set with 5 value options: -2, -1, 0, 1, 2. Positive values indicate positivity, excitement, or strong control, while negative values indicate negativity, lethargy, or feelings of inferiority and depression. The absolute value represents intensity. The 7 emotion labels are 3D points in the VAD 3D space, as shown in Fig. 1, whose coordinates are listed in TABLE I.

TABLE I. 7 Basic Emotions' Positions in VAD 3D Space

| 7 basic emotions' 3D coordinates in VAD space | | | | | | |
|---|---|---|---|---|---|---|
| **Happy** | **Sad** | **Surprise** | **Angry** | **Disgust** | **Fear** | **Neutral** |
| x | 1.7 | -1.3 | -1.6 | -2 | -1.8 | -2 | 0 |
| y | 1.8 | -1.5 | 1.5 | 1.2 | 1.2 | 0.5 | 0 |
| z | 1.5 | -1.4 | -0.5 | -1 | 1 | -2 | 0 |

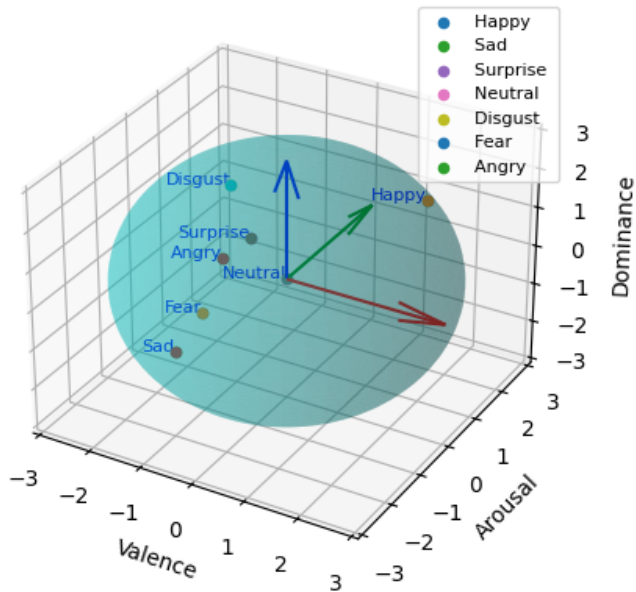

Fig. 1. 3D VAD space and 7 emotion positions in the space

Before the annotation process, training was provided to the annotation team, explaining the basic concepts of emotion recognition, the correlation between existing class labels in the FER2013 dataset and the VAD three-dimensional space, the meanings of each concept in VAD, and their extended implications. This was done to ensure that annotators have a deeper understanding of VAD during the annotation process, fostering better empathy for face images and enhancing annotation accuracy. Throughout the annotation process, annotators received continuous supervision, feedback, and timely guidance when faced with uncertain annotation samples.

*B. VAD annotation for FER2013*

The most popular FER datasets include FER2013[33], AffectNet [27], AFEW [34], and CK+ [35]. The last dataset of CK+ [35] is captured from laboratory where the participants act the specified expressions, all the other datasets of FER2013[33], AffectNet[27], and AFEW[34] could be regarded as in the wild. For smart online learning system, the students' facial expressions are captured naturally, which are consistent with the collection scenes that are in the wild, thus the former 3 datasets named FER2013[33], AffectNet[27], AFEW[34] could be used as the training dataset in the project. The annotation accuracies are different among them in which FER2013 is better than the other two. Additionally, although AffectNet[27] has dimensional annotations of Valence and Arousal, it does not label the Dominance parameter yet, which is as well the main work of the project.

There are two points that need clarification. First, not all images were included in the statistics because the remaining images did not meet the requirements for consistency. Therefore, they were not used. Second, the FER2013 dataset itself contains some erroneous images, i.e., images without faces. Such images were directly removed from this dataset. The dataset presented in this paper has a relatively high annotation accuracy and does not contain any erroneous images.

FER2013[33] is selected as the dataset in the research. It contains 35,887 face crops which are split into three parts, with 28709 images for training the orthogonal convolutional neural network, 3589 images for public test, and 3,589 images for private test. Current categorial expression labels include happy, sad, anger, disgust, surprise, fear and neutral. All images are grayscale and have a resolution of 48 by 48 pixels. In the project, it will be annotated in multidimensional space of Valence, Arousal, and Dominance as well.

Up to this time, it uses 18260 images in the FER2013, in which the number of images in each dataset are shown in TABLE II.

TABLE II. VAD ANNOTATED DATASET (FOR FER2013)

| Dataset Type | Image Number |
| --- | --- |
| Train | 14902 |
| Public Test | 1772 |
| Private Test | 1588 |
| Total | 18262 |

*C. VAD Annotation statistics analysis*

Fig.2 reveals the distribution of the entire dataset (including train, public test, private test) across values [-2, -1, 0, 1, 2] in Valence, Arousal, and Dominance. Overall observations indicate a tendency towards neutrality to negativity in Valence, positivity in Arousal, and a relatively balanced distribution with a slight inclination towards positivity in Dominance. From this chart, it can be inferred that, based on the analysis of over 7000 images, the overall emotions of these individuals are not very positive, but they are relatively excited and energetic, displaying a certain level of confidence and control.

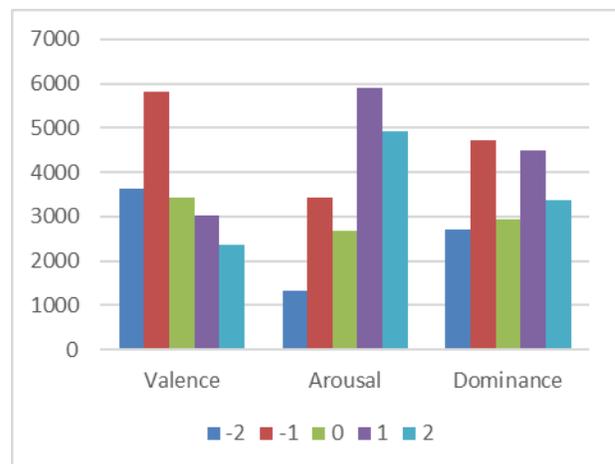

Fig. 2. VAD count for the annotated FER dataset

## IV. VAD Prediction Baseline

### A. Network Backbone

In this section, it introduces the orthogonalized regularization [37] to achieve more diverse and expressive features. First, it uses three independent regression networks to train and predict the VAD values respectively to avoid the correlation between VAD parameters affecting the training results. Then, each of them is based on ResNet-18 consisted of totally 17 convolution layers and 1 fully connected layer. Next, the input of each network is images and annotated VAD values ranged from -2 to +2, and the output is the generated VAD values. And finally, to achieve higher prediction performance, it extracts more diverse and expressive features by enforcing orthogonal convolution on several convolution layers in each network. Thus, the overall objective loss function is composed of the normal regression loss and the orthogonal convolution loss, which is presented in the following parts.

### B. Orthogonal Regularized Convolution

The orthogonality constraint on convolution layers [37] is developed from kernel orthogonality. The input image tensor $X$ is flattened into a column vector $X \in R^{C \times H \times W}$, the convolution kernel of each layer is reconstructed into DBT matrix[41] $K$, $K \in R^{MH'W' \times CHW}$ and thus the inner product of each row in $K$ and column in $X$ is equivalent to one convolution operation whose output is $Y$, $Y \in R^{MH'W'}$.

The orthogonal convolution corresponds to (1) or (2):

$$KK^T = I \text{ (Row Orthogonality).} \quad (1)$$

$$K^TK = I \text{ (Column Orthogonality)} \quad (2)$$

And thus, their orthogonalized regularization constraints are formulated as (3) or (4):

$$L_{korth-row} = \|KK^T - I\|_F \quad (3)$$

$$L_{korth-column} = \|K^TK - I\|_F \quad (4)$$

For output tensor $Y = Conv(X, K)$, $Y \in R^{MH'W'}$, the orthogonalized convolution constraints could be further reduced to (5):

$$Conv(K, K, padding = P, stride = S) = I_{r0} \quad (5)$$

Finally, it generates a near-orthogonal convolution regularization loss by (6):

$$L_{orth} = \|Conv(K, K, padding = P, stride = S) - I_{r0}\|_F^2 \quad (6)$$

To obtain orthogonalized neural network, L_orth is minimized by (7), which is the overall orthogonal regularization formulation.

$$\min_K \|Conv(K, K, padding = P, stride = S) - I_{r0}\|_F^2 \quad (7)$$

The overall objective loss function is obtained by combining the normal loss and orthogonal convolution regularization loss. Denoting λ>0 as the weight of orthogonal regularization loss, then the overall loss function is (8), where L_task is the normal loss of FER task, and L_orth is the orthogonal regularization loss.

$$L = L_{task} + \lambda L_{orth} \quad (8)$$

## V. Experiments

### A. Experiment Configuration

Firstly, the hardware of the system includes Intel(R) Xeon(R)E5-2678v3 CPU, 64G RAM, and NVidia 3090ti GPU with 24G memories. Secondly, the operating system is Windows10, GPU driver version is 537.13, CUDA version is 11.8, CUDNN version is 8.2.0.53, and the programming platform is Pytorch and Pycharm.

The orthogonal convolutional neural network is trained by MSE (Mean Squared Error) loss function and stochastic gradient descent with batch size of 64, because the output VAD are real continuous values. The learning rate is set to 0.01 initially and is reduced by a factor of 10 every 10,000 iterations. The total epochs are 120. Under the configurations, the training time is around 2 hours, and the experiment results are shown in the following part.

### B. Ablation Test & Result

The performance of the proposed orthogonal convolution neural network-based VAD prediction is compared with which without orthogonalized regularization. Their results are listed from Table 3 to Table 5. It is evident that on the annotated dataset, except for the results of the Dominance dimension in the public test, the ResNet-18 regression model with orthogonalization constraints on the Valence dimension, Arousal dimension, and Dominance in the private test (final test) demonstrates better predictive accuracy.

TABLE III.　'VALENCE' PREDICTION RESULTS (RMSE LOSS)

| Method | Public Test | Private Test |
|---|---|---|
| **Common Resnet-18+regression** | 0.076 | 0.063 |
| **Orthogonal convolution regularization** | **0.071** | **0.059** |

TABLE IV.　'AROUSAL' PREDICTION RESULTS (RMSE LOSS)

| Method | Public Test | Private Test |
|---|---|---|
| **Common Resnet-18+regression** | 0.048 | 0.094 |
| **Orthogonal convolution regularization** | **0.045** | **0.087** |

TABLE V.　TABLE 5 'DOMINANCE' PREDICTION RESULTS (RMSE LOSS)

| Method | Public Test | Private Test |
|---|---|---|
| **Common Resnet-18+regression** | 0.078 | 0.069 |
| **Orthogonal convolution regularization** | 0.080 | **0.066** |

Combining the experimental results of emotion classification in the [37], in which higher classification accuracy is obtained after orthogonalization, it could be concluded that, in the process of training convolutional neural networks for classification or regression, applying orthogonal regularization constraints to certain layers can enhance the network's capabilities and achieve better computational efficiency and performance.

*C. Prediction Disparity of VAD(Valence, Arousal and Dominance)*

The prediction loss in different dimensions of VAD is shown in Fig. 3. It is obvious that in the public test, A (Arousal) dimension achieves the highest accuracy, following which is the V (Valence) dimension that ranks second in accuracy, and D (Dominance) dimension has the lowest. On the other hand, in private test, V (Valence) reaches the first accuracy, D (Dominance) arrives the second, and A (Arousal) is in the third place.

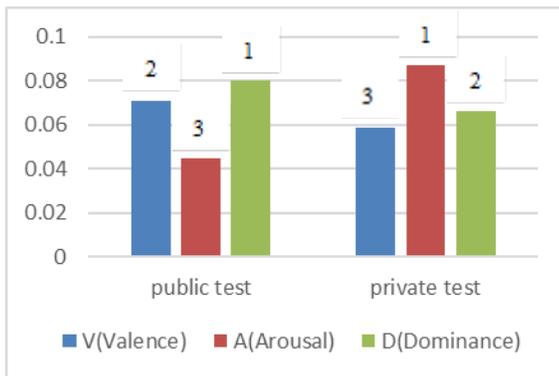

Fig. 3. Prediction Disparity of VAD and their ranks

After grading and summarizing their accuracy rank from both of public and private test, it results that V (Valence) ranks at the first place with 3 points, A (Arousal) is the second with 4 points, and D (Dominance) locates at the third place with 5 points, which represents that the prediction performance of V (Valence) is the best one, while D (Dominance) is the worst one. In fact, this result is consistent with the found the V dimension is the easiest to recognition, A Dimension is relatively more difficult to judge than V, and D dimension is the most challenging out of the three dimensions.

The meaning of D (Dominance), which represents dominance, was extended to include concepts of authority, confidence, and power to make it easier for annotators to understand. This broader interpretation, however, still makes it inconvenient to judge human's power and dominance based on the facial expression images. In fact, as far as the researchers of this study know, there is currently no facial expression dataset that includes annotations for the D dimension, which is the main objective of the study, i.e., annotating D dimension according to facial expression images.

With information available for all three VAD dimensions from facial expression images, obtained emotion information will be more rich, diverse, and comprehensive. This holds significant importance for future affective computing and intelligent emotion interactions in educational background.

## VI. CONCLUSION

Current systems almost exclusively use discrete emotion categories as of anger, disgust, tension, joy, sadness, surprise, and neutrality for quantification [12], but overlook the analysis from Valence (V), Arousal (A), Dominance (D) multidimensional space. However, future intelligent education system requires emotion quantification from more comprehensive and expressive perspectives. Therefore, current emotion quantification methods are challenging to meet the advanced requirements of emotion evaluation in intelligent emotion interaction.

Although researchers have developed emotion labelling in the two-dimensional space of valence (V) and arousal (A) in continuous form. However, VA alone cannot fully express emotion conditions. Thus, it proposes to add dominance (D, dominance-submissiveness) metric in this research, which can effectively explain human emotions, comprehensively quantify them by adopting the VAD three-dimension emotion metric.

It contributes to FER (facial expression recognition) in these three aspects: First, it introduces D(Dominance) dimension metric to measure emotions on FER dataset, which could be a benchmark for VAD emotion recognition. Then, it proposes a better regression network for VAD calculation through orthogonalized convolution, which could be the baseline for obtaining VAD multidimensional parameters from facial expression images. Finally, it provides a VAD (Valence, Arousal and Dominance) recognition tool of FER (Facial Expression Recognition) in future intelligent online education system.

ACKNOWLEDGMENT

R. B. G. thanks the students who annotate VAD values on FER2013 dataset, include but not limited to Sun Pengjun, Guo Zixin, Tang Xinyan, Zhang Yafei, Fu Runze, He Yihan, Chen Yunze, Wu Yuetong, Wang Yaxuan, Zhang Jie, Chen Zijian, Guo Weijie, Cui Haoran, Bian Liwen, Yuchi, Wu Fan, and Guo Weijie.

REFERENCES
[1] Yang Lu. Artificial intelligence: a survey on evolution, models, applications and future trends. Journal of Management Analytics. Volume 6, Issue 1, Feb, 2019.
[2] Shapsough, Salsabeel Y.; Zualkernan, Imran A., A Generic IoT Architecture for Ubiquitous Context-Aware Learning, Ieee
[3] Amponsah, S and Bekele, TA. Exploring strategies for including visually impaired students in online learning. (Early Access). Education and Information Technologies. Jun 2022.
[4] Clark A E , Nong H , Zhu H , et al. Compensating for academic loss: Online learning and student performance during the COVID-19 pandemic[J]. Post-Print, 2021.
[5] Wolf, M.A., Pizanis, A., Fischer, G. et al. COVID-19: a catalyst for the digitization of surgical teaching at a German University Hospital. BMC Med Educ 22, 308 (2022).
[6] Feuerlicht, G.; Beranek, M.; Kovar, V. Impact of COVID-19 pandemic on Higher Education. 2021 International Conference on Computational Science and Computational Intelligence, CSCI 2021: 1095-1098, 2021.
[7] Stoehr F , Mueller L , Brady A P , et al. How COVID-19 kick-started online learning in medical education—The DigiMed study[J]. PLoS ONE, 2021.

[8] Wolf, M.A., Pizanis, A., Fischer, G. et al. COVID-19: a catalyst for the digitization of surgical teaching at a German University Hospital. BMC Med Educ 22, 308 (2022).

[9] Addimando L. Distance Learning in Pandemic Age: Lessons from a (No Longer) Emergency. Int J Environ Res Public Health. 2022 Dec 5;19(23):16302.

[10] Zhao S , Yao H , Gao Y , et al. Predicting Personalized Image Emotion Perceptions in Social Networks[J]. IEEE Transactions on Affective Computing, 1949:1-1.

[11] Wang R , Shi Z . Personalized Online Education Learning Strategies Based on Transfer Learning Emotion Classification Model[J]. Hindawi, 2021.

[12] Mohamad Nezami, O., Dras, M., Hamey, L., Richards, D., Wan, S., Paris, C. (2020). Automatic Recognition of Student Engagement Using Deep Learning and Facial Expression. In: Brefeld, U., Fromont, E., Hotho, A., Knobbe, A., Maathuis, M., Robardet, C. (eds) Machine Learning and Knowledge Discovery in Databases. ECML PKDD 2019.

[13] Waleed Maqableh, Faisal Y. Alzyoud, Jamal Zraqou. The use of facial expressions in measuring students' interaction with distance learning environments during the COVID-19 crisis. Visual Informatics, Volume 7, Issue 1, 2023, Pages 1-17.

[14] Savchenko, AV; Savchenko, LV and Makarov, I. Classifying Emotions and Engagement in Online Learning Based on a Single Facial Expression Recognition Neural Network. IEEE Transactions on Affective Computing. 2022.

[15] Xiao J, Jiang Z, Wang L, Yu T. What can multimodal data tell us about online synchronous training: Learning outcomes and engagement of in-service teachers. Front Psychol. 2023 Jan 6;13:1092848.

[16] Mollahosseini A, Hasani B, Mahoor M H. AffectNet: A Database for Facial Expression, Valence, and Arousal Computing in the Wild[J]. IEEE Transactions on Affective Computing, 1949:1-1.

[17] Mehrabian A, Russell J A. An approach to environmental psychology[M]. MIT, 1974.

[18] Mehrabian A. Basic Dimensions for a General Psychological Theory. 1980.

[19] Ayvaz U, Use of Facial Emotion Recognition in E-learning Systems. 2017.

[20] Marvin Minsky. The Society of Mind. Simon & Schuster. March 15, 1988.

[21] Rosalind W. Picard. Affective computing. The MIT Press; Reprint edition, July 31, 2000.

[22] Pekrun Reinhard. Progress and open problems in educational emotion research [J]. Learning & Instruction, 2005, 15(5): 497-506.

[23] Zhao Shuyuan. Study on the Academic Emotions of College Students Based on the Control-Value Theory [D]. Central South University, 2013.

[24] Li L A , Yz A , Myc A , et al. Spontaneous facial expression database of learners' academic emotions in online learning with hand occlusion - ScienceDirect[J],2019.

[25] Mehrabian, Albert, and James A. Russell. An approach to environmental psychology. the MIT Press, 1974.

[26] Mehrabian, Albert (1980). Basic dimensions for a general psychological theory. pp. 39–53. ISBN 978-0-89946-004-8

[27] A. Mollahosseini, B. Hasani and M. H. Mahoor, "AffectNet: A Database for Facial Expression, Valence, and Arousal Computing in the Wild," in IEEE Transactions on Affective Computing, vol. 10, no. 1, pp. 18-31,1 Jan.-March 2019.

[28] Zhou Feiyan, Jin Linpeng, Dong Jun. A Comprehensive Review of Convolutional Neural Networks. Journal: Journal of Computer Science. January 2017.

[29] Zhang, Wei, et al. Learning a facial expression embedding disentangled from identity. Proceedings of the IEEE/CVF Conference on Computer Vision and Pattern Recognition. 2021.

[30] Gong S , Liu X , Jain A K . Mitigating Face Recognition Bias via Group Adaptive Classifier[J]. arXiv e-prints, 2020.

[31] Xie, Di, Jiang Xiong, and Shiliang Pu. All you need is beyond a good init: Exploring better solution for training extremely deep convolutional neural networks with orthonormality and modulation. Proceedings of the IEEE Conference on Computer Vision and Pattern Recognition. 2017.

[32] Wang J , Chen Y , Chakraborty R , et al. Orthogonal Convolutional Neural Networks[J]. 2019

[33] Goodfellow et al. Challenges in Representation Learning: A report on three machine learning contests. 1 Jul 2013, ICML 2013 Workshop.

[34] Abhinav Dhall, Roland Goecke, Simon Lucey, Tom Gedeon, Collecting Large, Richly Annotated Facial-Expression Databases from Movies, IEEE Multimedia 2012.

[35] P. Lucey, J. F. Cohn, T. Kanade, J. Saragih, Z. Ambadar and I. Matthews, "The Extended Cohn-Kanade Dataset (CK+): A complete dataset for action unit and emotion-specified expression," 2010 IEEE Computer Society Conference on Computer Vision and Pattern Recognition - Workshops, 2010.

[36] Pramerdorfer C , Kampel M . Facial Expression Recognition using Convolutional Neural Networks: State of the Art: 10.48550/arXiv.1612.02903[P]. 2016.

[37] Yi Huo, Lei Zhang. OCFER-Net: Recognizing Facial Expression in Online Learning System. AVI 2022: Proceedings of the 2022 International Conference on Advanced Visual InterfacesJune 2022